\documentclass[lettersize,journal]{IEEEtran}
\usepackage{amsmath,amsfonts}
\usepackage{array}
\usepackage[caption=false,font=normalsize,labelfont=sf,textfont=sf]{subfig}
\usepackage{textcomp}
\usepackage{stfloats}
\usepackage{url}
\usepackage{verbatim}
\usepackage{graphicx}
\usepackage{cite}
\usepackage{diagbox}
\usepackage{makecell}

\usepackage{amsmath,amssymb,amsfonts}
\usepackage{algpseudocode}
\usepackage{booktabs}

\usepackage{xcolor}
\usepackage{url} 
\usepackage{subfig}
\usepackage[ruled]{algorithm2e}
\usepackage{multirow}

\begin{document}

\title{Transaction Fraud Detection via Spatial-Temporal-Aware Graph Transformer}

\author{Yue~Tian,
        Guanjun~Liu,~\IEEEmembership{Senior Member,~IEEE,}
\thanks{This work has been submitted to the IEEE for possible publication. Copyright may be transferred without notice, after which this version may no longer be accessible.}
\thanks{Yue Tian and Guanjun Liu are with Department of Computer Science, Tongji University, Shanghai 201804, China (e-mail: 1810861@tongji.edu.cn; liuguanjun@tongji.edu.cn).}

}

\markboth{Journal of \LaTeX\ Class Files,~Vol.~14, No.~8, August~2021}%
{Shell \MakeLowercase{\textit{et al.}}: A Sample Article Using IEEEtran.cls for IEEE Journals}


\maketitle

\begin{abstract}
How to obtain informative representations of transactions and then perform the identification of fraudulent transactions is a crucial part of ensuring financial security. Recent studies apply Graph Neural Networks (GNNs) to the transaction fraud detection problem. Nevertheless, they encounter challenges in effectively learning spatial-temporal information due to structural limitations. Moreover, few prior GNN-based detectors have recognized the significance of incorporating global information, which encompasses similar behavioral patterns and offers valuable insights for discriminative representation learning. Therefore, we propose a novel heterogeneous graph neural network called Spatial-Temporal-Aware Graph Transformer (STA-GT) for transaction fraud detection problems. Specifically, we design a temporal encoding strategy to capture temporal dependencies and incorporate it into the graph neural network framework, enhancing spatial-temporal information modeling and improving expressive ability. Furthermore, we introduce a transformer module to learn local and global information. Pairwise node-node interactions overcome the limitation of the GNN structure and build up the interactions with the target node and long-distance ones. Experimental results on two financial datasets compared to general GNN models and GNN-based fraud detectors demonstrate that our proposed method STA-GT is effective on the transaction fraud detection task.
\end{abstract}


\section{Introduction}
\label{sec:introduction}
Transaction fraud incidents frequently occur in the rapidly evolving development of financial services, leading to substantial economic losses \cite{ref21}. According to the Nielsen report, global credit card losses amounted to 25 billion dollars in 2018, and further increases are expected \cite{ref22}. 
Consequently, identifying fraudulent transactions is crucial to mitigate financial losses, enhance customer experience, and safeguard the reputation of financial institutions.

Numerous techniques have been proposed for detecting transaction fraud, and they can be classified into two categories: rule-based methods and machine learning-based methods.
1) The rule-based methods rely on human-designed rules with expert knowledge to assess the likelihood that fraud has occurred \cite{ref23}. These methods heavily rely on experts' domain knowledge which cannot perform well in complex environments.
Moreover, the fixed rules limit the algorithm's ability to adapt to dynamic fraud patterns.
2) The machine learning-based methods can detect fraudulent transactions automatically by constructing supervised or unsupervised models leveraging vast historical transaction data \cite{ref25}. Machine learning-based methods usually resort to feature extraction \cite{ref26}. Achieving statistical features from transaction attributes is feasible such as time, location, and amount. However, incorporating unstructured data such as device ID and WiFi position is challenging to extract. Additionally, effectively capturing the interaction between transactions presents difficulties. Therefore, applying machine learning-based methods to identify fraud is still constrained by itself.

Graph-based approaches have recently exhibited superior performance in fraud detection \cite{ref13, ref16, ref19}.
GNN techniques acquire the representation of the central node through the selective aggregation of information from neighboring nodes \cite{ref29, ref30}. In contrast to conventional fraud detection methods, they can facilitate automatic feature learning by capturing the interactive relationships between transactions. Additionally, graph-based approaches can efficiently identify fraudulent transactions through end-to-end learning \cite{ref37}.

However, graph-based fraud detection methods encounter significant challenges when faced with the following problems.
First, applying the GNN method for our fraud detection task needs to pay attention to the learning of spatial-temporal information. We have the following observations for fraudulent transactions:  1) Spatial aggregation: Fraudsters often utilize a limited number of devices to execute fraudulent activities, as acquiring transaction equipment incurs costs. 2) Temporal aggregation: Fraudulent actions are frequently undertaken within a narrow time frame, as the detection of suspicious behavior by the cardholder or financial institution can prompt the termination of the transaction. Some recent works, including GEM \cite{ref31} and STAGN \cite{ref16}, have noted similar challenges. GEM establishes a connection with the account that occurred on the device within the same time period \cite{ref31}. STAGN leverages temporal and spatial slices to consider both spatial and temporal aggregation \cite{ref16}. However, they fail to distinguish the temporal differences of neighbor transactions in the same time slice, as shown in Fig.~1(a). Meanwhile, the representation of the target transaction may depend on the ones in other time slices, as shown in Figs.~1(b) and (c). In this way, due to structural limitations, informative transactions that satisfy the homogeneity assumption cannot be fully exploited. Therefore, it is unreasonable to construct a separate transaction graph for each time slice and then perform graph convolution operation to incorporate spatial-temporal information.

Secondly, graph-based detection methods rely on aggregating information from neighbor nodes to update the representation of target nodes. However, this approach only utilizes local information while ignoring global information. In fact, long-distance transactions may contain similar information. GNN-based fraud detectors cannot capture the information to obtain discriminative representations. For example, the target transaction $a$ needs to use the $K$-hop transaction $b$. Although we can obtain the information of transaction $b$ by stacking $K$ layers of GNN layers, it may cause a dilution of information from transaction $b$. Simply expanding the receptive field of GNN is insufficient for learning discriminative representations. With the depth increases, GNNs may face the over-smoothing issue, where the learned representations of each node tend to become consistent. It results in the limited expressive ability for GNN-based fraud detectors.

\begin{figure*}[!t]
\centering
\includegraphics[width=5.2in]{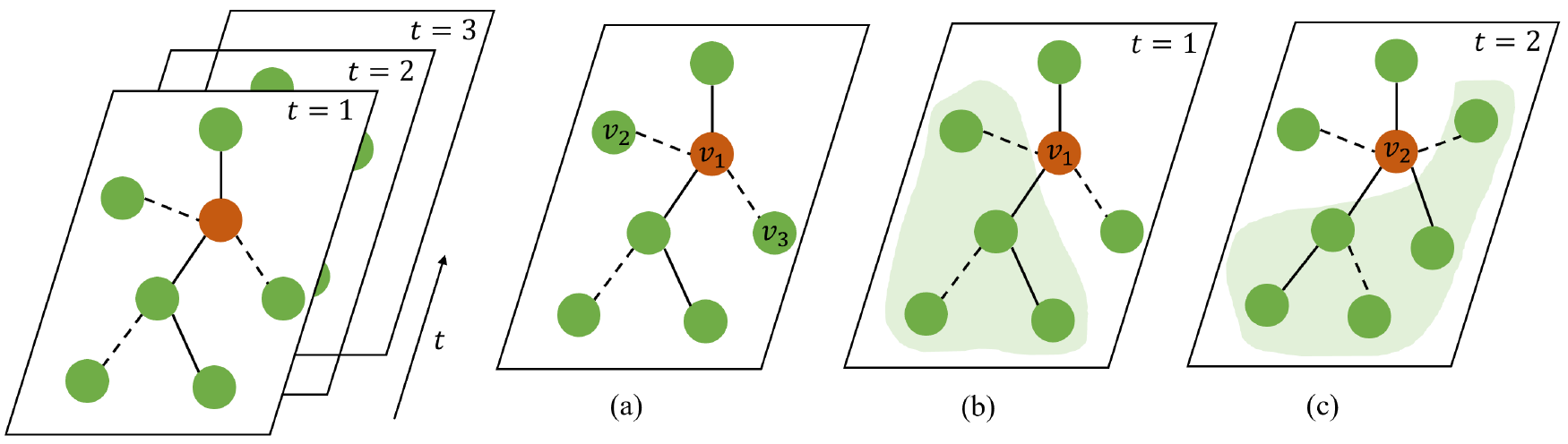}
\caption{The general transaction graph taking transactions as nodes. As shown in Fig.~1(a), given the target node $v_1$, the general GNNs in fraud detection tasks cannot distinguish temporal differences between nodes $v_2$ and $v_3$. Nodes $v_1$ in Fig.~1(b) and $v_2$ in Fig.~1(c) have the same attributes but are located in different time slices. The general GNNs cannot exploit informative nodes due to structural limitations and the over-smoothing issue.}
\label{fig_1}
\end{figure*}

To address the aforementioned challenges, we propose a novel graph neural network model to detect fraudulent transactions, called \emph{Spatial-Temporal-Aware Graph Transformer} (STA-GT). First, STA-GT is built on a heterogeneous graph neural network to model spatial-temporal information for learning discriminative representations. Specifically, a heterogeneous graph is constructed, which takes transactions as nodes and consists of various edge types (e.g., IP address and MAC address) based on the transaction location. 
To capture temporal dependencies, we incorporate a designed temporal encoding strategy into the graph neural network architecture, which makes STA-GT gather spatial-temporal information effectively.
To further improve STA-GT's performance, we leverage a relation-level attention mechanism to  specify the contributions of different relations dynamically and concatenate the intermediate embeddings from the corresponding GNN layers to deal with varying degrees of sharpness and smoothness.
Finally, a Transformer sub-network is added on top of the heterogeneous GNN layer stack. In this way, STA-GT incorporates global information into its learning process while preserving the GNN's ability to capture local structural information.
Extensive experiments are conducted on two financial datasets to evaluate the performance of STA-GT. Compared to other state-of-art methods, the experimental results demonstrate its superiority in fraud detection tasks.

The contributions of this paper are summarized as follows:

\begin{enumerate}[]
    \item We propose a heterogeneous graph neural network method to identify fraudulent transactions. It can learn spatial-temporal information while preserving structural information.
    To the best of our knowledge, it is the first work that employs a graph neural network integrated with the temporal encoding strategy to model spatial-temporal dependencies on the transaction fraud problem.
     \item We overcome the limitation of the GNN structure to propose a local-global learning module, which can capture all pairwise node-node interactions and build up the connections between the target node and long-distance neighbors. By incorporating this module, STA-GT is able to effectively learn both local and global transaction information while alleviating the over-smoothing phenomenon.
     \item We construct experiments on two financial datasets, including performance comparison, ablation studies, and parameter sensitivity analysis. The results show that STA-GT outperforms other baselines on the transaction fraud detection task.
\end{enumerate}

The rest of the paper is organized as follows. Section II presents the related work. Section III introduces the problem definition for the transaction fraud tasks. Section IV describes how the STA-GT identifies fraudulent transactions. Section V introduces the datasets and evaluates the performance of STA-GT compared with the other GNN-based baselines. Section VI concludes the paper.

\section{RELATED WORK}
\subsection{Graph Neural Networks}
The GNNs' excellent ability to process non-structured data has made them widely applied in electronic transactions, recommendation systems, and traffic forecasting \cite{ref39,ref40}. Its basic idea is to obtain the representation of each node by leveraging the information from itself as well as its neighboring nodes. GNNs are divided into two categories. 1) Spectral neural networks propose graph convolution operations in the spectral domain.
ChebNet approximates graph convolution using polynomial expansion \cite{ref9}.
GCN performs spectral convolutions on graphs to capture structure and feature information \cite{ref1}. 2) Spatial Graph Neural Networks apply convolution operations on the graph structure through leveraging the information of neighborhood nodes. GraphSAGE proposes a general inductive framework, which can efficiently update the representation of the target node \cite{ref2}. It utilizes the defined aggregators to sample and aggregate local neighborhood information of the target nodes \cite{ref2}. GAT leverages a self-attention mechanism to enable distinct treatment of various neighbors during the embedding updating of the target node \cite{ref3}.

To model heterogeneity and learn rich information, heterogeneous graph neural networks are proposed. RGCN is an extension method of GCN to model the relational data \cite{ref6}. HAN utilizes a hierarchical attention strategy to evaluate the corresponding significance of neighbors and meta-paths. According to the learned importance, HAN can learn the complex structure and feature information to generate the representations of each node \cite{ref7}. 

However, these methods are not explicitly designed for our transaction fraud detection task. And they ignore the problem of temporal-spatial dependency and how to make full use of informative but long-distance transactions.

\subsection{GNN-based Fraud Detection}
Recently, some researchers have explored how to apply GNNs to fraud detection tasks, revealing the suspiciousness of fraudulent behaviors. Based on various scenarios, GNN-based fraud detection is divided into two categories: financial fraud detection \cite{ref13, ref16, ref19} and opinion fraud detection \cite{ref14, ref15, ref17,ref18}. GEM is the pioneering work to detect malicious accounts via a heterogeneous graph neural network \cite{ref13}. CARE-GNN designs a label-aware similarity sampler with a reinforcement learning strategy to solve two camouflage issues, including the feature and relation camouflage \cite{ref14}. To address the issue of imbalanced node classification, PC-GNN introduces a label-balanced sampler for reconstructing sub-graphs\cite{ref17}. It employs an over-sampling technique for the neighbors belonging to the minority class and an down-sampling technique for the others \cite{ref17}. To handle feature inconsistency and topology inconsistency, FRAUDRE integrates several key components, including the topology-agnostic embedding layer, the fraud-aware graph operation, and the inter-layer embedding fusion module \cite{ref15}. Moreover, to mitigate the impact of class imbalance, the imbalance-oriented loss function is introduced \cite{ref15}.
STAGN aims to learn spatial-temporal information via an attention-based 3D convolution neural network \cite{ref16}. MAFI alleviates the camouflage issue via a trainable sampler and utilizes the relation-level and aggregator-level attention mechanisms to specify the corresponding contributions \cite{ref18}.
xFraud adopts a self-attentive heterogeneous graph neural network to automatically aggregate information from different types of nodes without predefined meta-paths and designs a hybrid explainer which is a tradeoff between GNN-based explanations and traditional topological measures \cite{ref19}.

Among these methods, only two works \cite{ref13, ref17} noticed the temporal information. While \cite{ref13} only captures the interaction between two nodes that occurred on the device within the same time period, and \cite{ref17} utilizes the temporal slices. They ignore the spatial-temporal information in other time slices. STA-GT remedies the shortcoming via the temporal encoding strategy. Furthermore, the above methods fail to use global transaction information for great expressive ability.
\begin{figure*}[!t]
\centering
\includegraphics[width=6in]{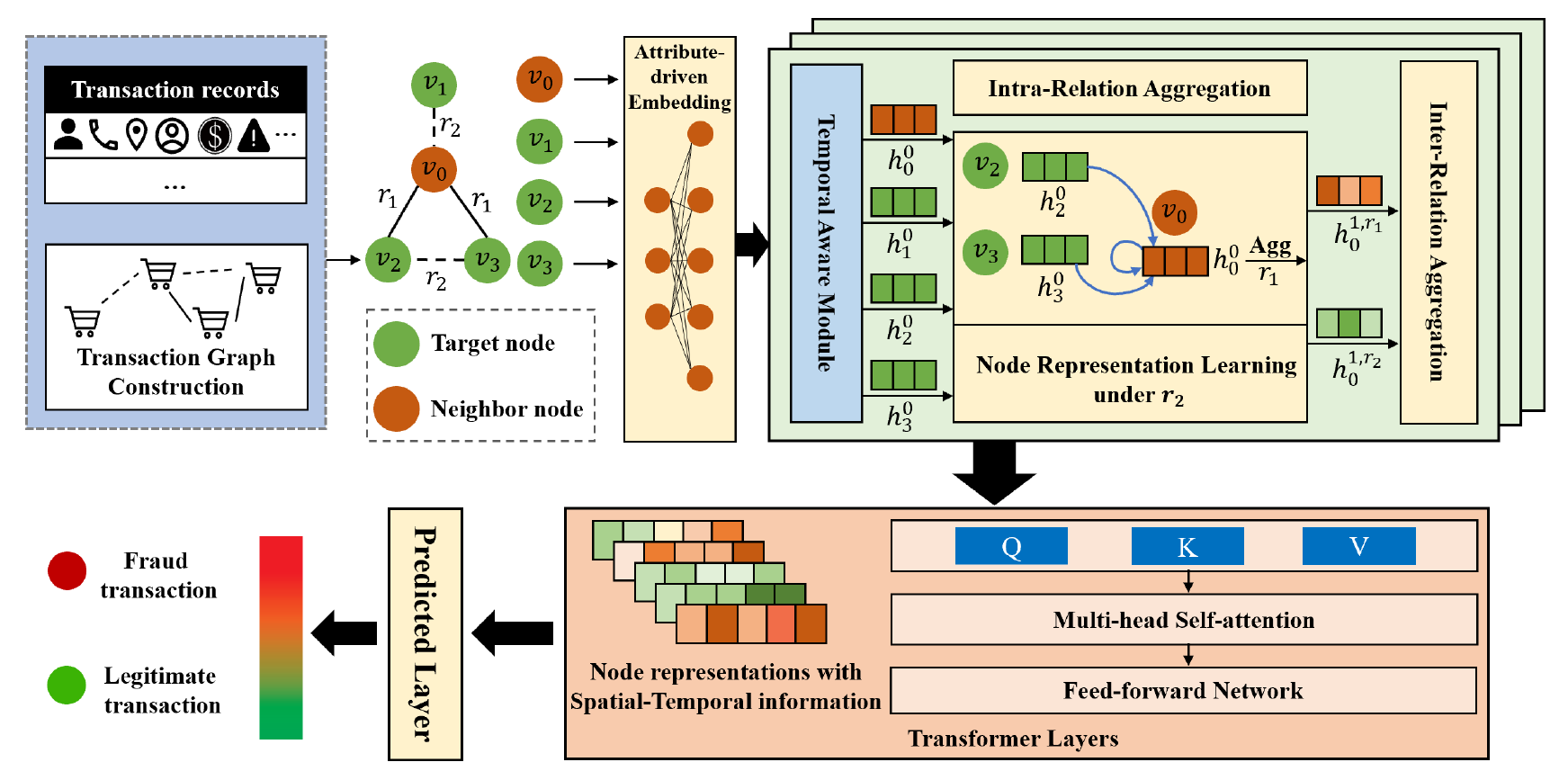}
\caption{The framework of STA-GT. The topology-agnostic dense layer utilizes the original attributions to learn representations for all nodes. Followed by the temporal-aware module, the temporal information of all nodes can be captured. All representations containing temporal encoding are inputted to the graph convolution module for next updating. The intra-relation aggregation, inter-relation aggregation, and intermediate representation combination are performed. After modeling the spatial-temporal dependencies, the transformer layer learns global information to connect the target node $v_0$ and the long-distance nodes. Finally, the target node $v_0$ is classified by the predicted layer.
}
\label{fig_1}
\end{figure*}

\section{PROBLEM DEFINITION}

In this section, we present the conceptions of multi-relation graph. Then, we formulate fraud detection on the graph problem.


\noindent \textbf{Definition 1. \emph{Multi-relation Graph.}}

A multi-relation graph is defined as $\mathcal{G} = \{\mathcal{V},\{\mathcal{E}\}_{1}^{R},\mathcal{X},\mathcal{Y} \}$, where $\mathcal{V}$ and $\mathcal{E}$ respectively are the sets of nodes and edges. $e_{i,j}^{r}$ denotes an edge which connects nodes $i$ ans $j$ under the relation $r \in \{ 1,...,R \}$.  Each node refers to a transaction record $x$, where $x$ is a $d$-dimensional feature vetcor denoted as $x_i \in \mathcal{R}^d$ and the set of node features are represented as $\mathcal{X} = {x_1, ..., x_n}$. $\mathcal{Y}$ represents the set of labels of all nodes $\mathcal{V}$.

\noindent \textbf{Definition 2. \emph{Fraud Detection on the Graph.}}

In the transaction graph, each node $v$ denotes a transaction whose suspiciousness needs to be predicted. And its label is denoted as $y_v \in \{0,1\}$, where 0 and 1 represent legitimate and fraudulent transactions, respectively. The identification of fraudulent transactions is a semi-supervised binary classification problem. A transaction fraud detection model can be trained using information from labeled nodes. Then, it is utilized to infer the possibility that the unlabeled node is predicted to be fraudulent.

\section{PROPOSED MODEL}
In this section, the pipeline of our method STA-GT is introduced, as shown in Fig.~1. STA-GT has five modules: 1) \textbf{Attribute-driven Embedding}. The topology-agnostic layer is utilized to obtain initial layer embeddings. 2) \textbf{Temporal aware module.} The temporal encoding strategy is used to capture the temporal dependency. 3) \textbf{Aggregation process.} Intra-relation and inter-aggregation are performed to learn the embeddings in each layer with the full utilization of the information from the target node and its neighbors while keeping the contributions of different relations.
And then, we concatenate these intermediate embeddings to fuse all information. Spatial-Temporal information can be learned by the above step. 4) \textbf{Transformer layer.} Global information can be captured by making use of pair-wise node-node interactions. 5) \textbf{Predicted layer.} It is used to classify whether the transaction is fraudulent or not.

\subsection{Attribute-driven Embedding}
Fraudsters often mimic cardholders' behavior to avoid their suspicions, leading to fraudulent nodes and normal ones often exhibit similarities. Consequently, utilizing the original node features to learn representations is imperative before the GNN training. In this study, we employ the attribute-driven embedding layer \cite{ref15} to facilitate the learning of feature similarity without relying on graph topological information.
Given the node $v_i$, the initial layer embedding can be denoted as:

\begin{equation}
\begin{aligned}
h_{v_i}^{0}=\sigma(x_iW_1),
\end{aligned}
\end{equation}
where $\sigma$ represents a non-linear activation function, $xi \in \mathbb{R}^s$ denotes the original attributes of node $v_i$, and $W_1 in \mathbb{R}^{s \dot d}$ denotes a learnable weight matrix.

\subsection{Modeling the Temporal Dependency}


In the field of transaction fraud detection, the conventional approach  integrated temporal information usually construct separate graphs for each time slice.
However, they cannot distinguish the temporal dependency from different neighbor nodes and break structural limitations allowing the target node to interact with nodes in other time slices. In the domain of transaction fraud detection, the prevailing approach for incorporating temporal information involves generating separate graphs for each time slice. Nonetheless, this methodology fails to differentiate temporal dependencies from various neighboring nodes in the same time slice and 
cannot break structural limitations enabling interactions between the target node and nodes existing in different time slices. To capture temporal information and maximize the use of structural information, we allow the target node to interact with the nodes that occur at any time. And we define the temporal encoding strategy, allowing nodes to learn a hidden temporal representation. Given a target node $v_i$ at time $t(v_i)$, the temporal encoding can be expressed as follows:
\begin{equation}
\begin{aligned}
Base(t(v_i), 2i) = sin(\frac{ t_{v_i}}{10000^{\frac{2i}{d}}}),
\end{aligned}
\end{equation}

\begin{equation}
\begin{aligned}
Base(t(v_i), 2i+1) = cos(\frac{ t_{v_i}}{10000^{\frac{2i+1}{d}}}),
\end{aligned}
\end{equation}

\begin{equation}
\begin{aligned}
TE(t_{v_i})=T-Linear(Base(t(v_i))),
\end{aligned}
\end{equation}
where $T-Linear$ is a tunable linear projection. Then, we can model the relative temporal dependency between nodes $v_i$ and $v_j$. By adding the temporal encoding, the hidden representation of node $v_i$ can be updated as follows:

\begin{equation}
\begin{aligned}
h_{v_i}^{0, t} = h_{v_i}^{0} + TE(t_{v_i}).
\end{aligned}
\end{equation}

By adopting this approach, the enhanced temporal representation becomes capable of capturing the relative temporal relationships between the target node $v_i$ and its neighbor nodes.

\subsection{Modeling the Spatial Dependency}

The transaction graph $\mathcal{G}$ encodes the relationships among the transactions. The connected transactions in the $\mathcal{G}$ tend to share similar features. Specific to our fraud detection problem, fraudsters connect with others since they always leverage shared devices to execute fraudulent activities. Hence, we utilize the GNN method to model the spatial dependency. Given a node $v$ and its hidden embedding, which contains temporal information after the aforementioned step, we leverage the following intra-relation and inter-relation aggregation mechanisms to update its representation. Subsequently, by concatenating the intermediate layer embeddings, we obtain a comprehensive representation that incorporates both spatial and temporal patterns.

\subsubsection{Intra-Relation and Inter-Relation Aggregation}
~\\
Given a node $v_i$ and its neighbor node $v_j$ under $r$ relation at $\ell$-th layer, we learn the neighborhood information under the homophily assumption and the difference between them. The graph convolution operation is denoted as:
\begin{equation}
\begin{aligned}
h_{i,r}^{\ell,t}=&COMBINE(AGGR\{h_{v_j}^{\ell-1,t'}, v_j \in \mathcal{N}_r(v_i)\}, \\
&AGGR\{h_{v_i}^{\ell-1,t}-h_{v_j}^{\ell-1,t'}, v_j \in \mathcal{N}_r(v_i)\}),
\end{aligned}
\end{equation}
where $h_{v_j}^{\ell-1,t}$ denotes the $\ell$-th layer embedding with temporal information of node $v_i$ under the $r$-th relation, $t$ denotes the timestamp of node $v_i$, $t'$ denotes the timestamp of node $v_j$, $\ell \in \{1,2,...,L\}$, $r \in \{1,2,...,R\}$, and $\mathcal{N}_r(v_i)$ is the set of neighbors of node $v_i$ under relation $r$.

Considering that the different relations provide corresponding contributions, we employ the attention mechanism to specify the importance of each relation. The representation of node $v_i$ under $R$ relations is denoted as:

\begin{equation}
\begin{aligned}
h_{v_i}^{\ell,t}=\sum_{r=1}^{R} \alpha_{r}^{\ell} \odot h_{i,r}^{\ell,t},
\end{aligned}
\end{equation}
where $\alpha_{r}^{\ell}$ denotes the normalized importance of relation $r$. $\alpha_{r}^{\ell}$ can be formulated as follows:
\begin{equation}
\begin{aligned}
w_{r}^{\ell}=\frac{1}{|V|} \sum_{i \in \mathcal{V}} q^T \cdot tanh(W_2 \cdot h_{i,r}^{\ell,t}+b),
\end{aligned}
\end{equation}

\begin{equation}
\begin{aligned}
\alpha_{r}^{\ell}=\frac{exp(w_{r}^{\ell})}{\sum_{i=1}^{R}exp(w_{r}^{\ell})},
\end{aligned}
\end{equation}
where $q$, $W_2$, $b$ denote the relation level vector, the weighted matrix, and the bias vector, respectively. 

\subsubsection{Inter-layer Representation Fusion}
~\\
The node representations outputted of different layers in the GNN architecture manifest distinct levels of sharpness and smoothness information, as evidenced by prior studies \cite{ref15}. The initial layers of the network predominantly capture localized information, whereas the subsequent layers exhibit an increased ability to capture global information, as supported by previous research \cite{ref38}. Consequently, to obtain discriminative node embeddings, we concatenate the intermediate embeddings derived from the corresponding GNN layers:

\begin{equation}
\begin{aligned}
h_{v_i}^{t}=COMBINE(h_{v_i}^{1,t},h_{v_i}^{2,t},...,h_{v_i}^{L,t}).
\end{aligned}
\end{equation}

Following the above step, graph convolution operations are performed to capture the local neighbor information and the relative temporal relationships between nodes, thereby facilitating the modeling of spatial-temporal information.

\subsection{Learning Global Information}
Given the above spatial-temporal information learning, the next step is how to obtain global information for the target node which exhibits similar behavioral patterns. Inspired by \cite{ref33}, we use a transformer layer for each node individually. Specifically, a multi-head attention mechanism is performed on the above obtained embedding matrix that denotes $H^{v_i} \in \mathbb{R}^{N \times d}$, where $v_i$ represents the node $v_i$. Initially, we present the single-head attention approach, which is subsequently expanded to a multi-head attention mechanism. Firstly, we perform a linearly project on $H^{v_i}$ to obtain queries, keys, and values as follows:

\begin{equation}
\begin{aligned}
Q^{v_i}=H^{v_i}W^Q, K^{v_i}=H^{v_i}W^K, V^{v_i}=H^{v_i}W^V,
\end{aligned}
\end{equation}
where $W^Q$, $W^K$, $W^V$ are the trained projection matrices, respectively. Hence, the single-attention function is defined as:

\begin{equation}
\begin{aligned}
Attention(H^{v_i}) & =softmax(\frac{Q^{v_i}{K^{v_i}}^T}{\sqrt{d_k}}V^{v_i}) \\
& = softmax(\frac{(H^{v_i}W^Q)(H^{v_i}W^K)^T}{\sqrt{d_k}}H^{v_i}W^V,
\end{aligned}
\end{equation}

The multi-attention function can be expressed as the concatenation of the outputs from individual attention function:

\begin{equation}
\begin{aligned}
Multihead(H^{v_i}) = Concat(head_1,...,head_s)W^O,
\end{aligned}
\end{equation}

\begin{equation}
\begin{aligned}
head_s & = attention_s(H^{v_i}) \\
       & = softmax(\frac{(H^{v_i}W_{s}^{Q})(H^{v_i}W_{s}^{K})^T}{\sqrt{d_k}}H^{v_i}W_s^V),
\end{aligned}
\end{equation}
where $W_{s}^{Q}$, $W_{s}^{K}$, and $W_{s}^{V}$ are the projection matrices of the $s$-th attention head, respectively. $W^O$ denotes also a linear projection. Subsequently, the output of the multi-head attention layer is passed through a point-wise feed-forward neural network, a residual layer, and a normalization layer. This sequential process ultimately leads to the update of the representations for all nodes, which are denoted as $H^{v_i}_{out} \in \mathbb{R}^{N \times d}$, such that local-global information can be captured.

\subsection{The Prediction Layer}
For each node $v$, we generate the final representation $h_{out}^{v_i}$ by integrating the above local spatial-temporal information and global information. Subsequently, the MLP classifier is employed to achieve the node classification task, that is, the identification of fraudulent transactions. The optimization of this process is carried out by the cross-entropy loss function \cite{ref34}.
\begin{equation}
\begin{aligned}
L = -\sum_{v\in V}{y_vlogP_v+(1-y_v)log(1-P_v)},
\end{aligned}
\end{equation}

\begin{equation}
\begin{aligned}
P_v= \sigma(MLP(z_v)),
\end{aligned}
\end{equation}
where $y_v$ denotes the real label of node $v$.

\section{Experiments}
In this section, we perform the experiments to investigate the superiority of the proposed method STA-GT on our transaction fraud detection tasks.

\subsection{Datasets and Graph Construction}

We conduct experiments on one private dataset and one public dataset to indicate that STA-GT achieves significant improvements compared to both classic methods and state-of-the-art GNN-based fraud detectors.

The private dataset, PR01, consists of 5.2 million transactions that took place in 2016 and 2017. Transactions are labeled by professional investigators of a Chinese bank, with 1 representing fraudulent transactions and 0 representing legitimate ones. In data pre-processing, we first utilize the down-sampling of legitimate transactions to solve the imbalanced problem. Then, we apply one-hot coding and min-max normalization to handle the discrete and continuous values, respectively. For our experimental setup, the training set comprises transactions from the first month, while the remaining transactions are partitioned into five distinct groups (PR1 to PR5) to serve as the test set. Transactions are represented as nodes, and there exist two relations among these nodes. Specifically, the $Trans-IP-Trans$ relation links transactions that occurred at the same IP address. The $Trans-MAC-Trans$ relation is employed to establish links between transactions that have occurred on the same MAC address.

The TC dataset\footnote{https://challenge.datacastle.cn/v3/} contains 160,764 transaction records collected by Orange Finance Company, including 44,982 fraudulent transactions and 115,782 legitimate transactions. We perform the same data processing as for the private dataset. The training set utilized in this study comprises transaction records from a designated week, while the subsequent week's transaction records constitute the test set. In this way, the TC dataset is split into TC12, TC23, and TC34. Transactions are also represented as nodes, and there exist four relations among these nodes: $Trans-IP-Trans$, $Trans-MAC-Trans$, $Trans-device1-Trans$, and $Trans-device2-Trans$.

\subsection{Baselines}
We compare the proposed method STA-GT with homogeneous GNNs (GCN, GraphSAGE, and GAT), heterogeneous GNNs (RGCN, HAN), and GNN-based fraud detectors (CARE-GNN, SSA, and FRAUDRE) to demonstrate its superiority. The baselines we choose are introduced as follows:

\begin{itemize}
    \item \textbf{GCN} \cite{ref1}: It is a traditional homogeneous GNN method that employs the efficient layer-wise propagation rule based on the first-order approximation of spectral convolutions.
    \item \textbf{GraphSAGE} \cite{ref2}: It is an inductive framework for learning node embedding from selective local information on the homogeneous graph.
    \item \textbf{GAT} \cite{ref3}: It is a homogeneous GNN method that leverages a self-attention strategy to specify the importance of neighbor nodes.
    \item \textbf{CARE-GNN} \cite{ref14}: It is a heterogeneous GNN architecture that effectively addresses the challenge of camouflages in the aggregation process of GNNs.
    \item \textbf{Similar-sample + attention SAGE (SSA)} \cite{ref5}: It is a GNN method performed on a multi-relation graph, which proposes a new sampling policy and a new attention mechanism to ensure the quality of neighborhood information.
    \item  \textbf{RGCN} \cite{ref6}: It is a GNN method designed to to address the challenges posed by complex, multi-relational data in tasks including entity classification and link prediction.
    \item \textbf{HAN} \cite{ref7}: It is a heterogeneous GNN method that employs hierarchical attention at both node and semantic levels, enabling the incorporation of the significance of both nodes and meta-paths. 
    \item \textbf{FRAUDRE} \cite{ref15}: It is a graph-based fraud detection framework to effectively tackle the challenges of imbalance and graph inconsistency. To handle feature inconsistency and topology inconsistency, the model integrates several key components, including the topology -agnostic embedding layer, the fraud-aware graph operation, and the inter-layer embedding fusion module. Moreover, to mitigate the impact of class imbalance, the imbalance-oriented loss function is introduced.
\end{itemize}

Note that the above methods, including GCN, GraphSAGE, GAT, and SSA, are applied to homogeneous graphs, treating each relation equally. CARE-GNN, RGCN, FRAUDRE, and HAN are used on multi-relation graphs to aggregate information under different relations.

\subsection{Evaluation Metrics}
To compare the performance of our approach STA-GT with the baseline models, $Recall$, $F1$, and $AUC$ are adopted as evaluation metrics. The metrics are briefly calculated as follows:
\begin{equation}
Recall= \frac{T_P}{T_P+T_N},
\end{equation}
where $T_P$, $T_N$, and $F_P$ are the numbers of true positive transaction records, true negative transaction records, and false positive transaction records, respectively \cite{ref36}.
\begin{equation}
Precision= \frac{T_P}{T_P+F_P},
\end{equation}
\begin{equation}
F_1= \frac{2 \times Recall \times Precision}{Recall+Precision},
\end{equation}

\begin{equation}
AUC= \frac{\sum_{r \in \mathcal{R^{+}}}rank_{r}- {\frac{|\mathcal{R^{+}}| \times (|\mathcal{R^{+}}|+1)}{2}} }{  |\mathcal{R^{+}}| \times |\mathcal{R^{-}}|  },
\end{equation}
where$\mathcal{R^{+}}$ and $\mathcal{R^{-}}$ are the fraudulent and legitimate class sets and $rank_{r}$ is the rank of $r$ by the predicted score.
For the mentioned metrics, a higher value indicates better model performance.

\begin{table*}[ht]
	\centering
	\caption{\centering{PERFORMANCE COMPARISON OF STA-GT AND ALL BASELINES ON THE PRIVATE DATASET.}}
	\label{tab:3}   
	\scalebox{1}{
	\begin{tabular}{l|c|ccccccccc}
		\toprule[2pt]
		\textbf{Dataset}     & \textbf{Criteria}     &  \textbf{GCN}    & \textbf{GraphSAGE} & \textbf{GAT}  &\textbf{CARE-GNN}   &\textbf{SSA}    &\textbf{RGCN} &\textbf{HAN}   & \textbf{FRAUDRE}  & \textbf{STA-GT}\\
            \midrule
            &               $Recall$(\%) & 62.0 & 60.5 & 69.2 & 79.9 & 63.6 & 66.6 & 57.6 &82.6& \textbf{86.4}\\
		\textbf{PR1}  &$F_1$(\%)    & 68.1 & 67.8 & 68.6 & 69.0 & 61.0 & 78.8 & 62.6 &75.8& \textbf{82.9}\\
		  &               $AUC$(\%)    &  87.6 & 87.7 & 86.5 & 91.1 & 82.6 & 83.0 & 75.2 &90.8& \textbf{92.9}\\
            \midrule
            &               $Recall$(\%) &  59.3& 68.4 & 75.0 & 86.2  & 67.9 & 62.1 & 51.8 &82.3& \textbf{87.3}\\
		\textbf{PR2}  &$F_1$(\%)    & 71.4 & 77.9 & 77.6 & 83.9 & 54.0 & 77.1 & 64.7 &75.1& \textbf{85.0}\\
		  &               $AUC$(\%)    & 85.2 & 87.9 & 86.5 & 94.1 & 87.4 & 79.4 &74.0  &91.5& \textbf{94.3}\\
    \midrule
          &               $Recall$(\%) & 81.7  & 70.2 & 82.1 & 85.3 & 73.2 & 61.8 & 63.9 &88.8& \textbf{90.5}\\
		\textbf{PR3}  &$F_1$(\%)    & 89.6 & 64.0 & 89.8 & 87.7 & 83.4 & 75.8 & 77.8 &90.6& \textbf{91.9}\\
		  &               $AUC$(\%)    & 86.9 & 83.2 & 87.1 & 98.2 & 79.8 & 79.2 & 81.5 &98.3& \textbf{98.4}\\
    \midrule
      &               $Recall$(\%) & 82.1 & 72.3 & 84.6 & 80.8 & 67.2 & 64.2 & 87.3 &88.3& \textbf{93.4}\\
		\textbf{PR4}  &$F_1$(\%)    & 89.9 & 79.3 & 91.3 & 70.7 & 79.9 &  77.2& 93.0 &90.8& \textbf{94.2}\\
		  &               $AUC$(\%)    & 85.6 & 88.2 & 87.5  & 91.5 &  73.7&  81.8 & 92.6 &\textbf{98.3}& 98.0\\
    \midrule
      &               $Recall$(\%) & 82.9 & 77.5 & 86.1 &82.1  & 52.4 & 66.1 & 85.6 &84.3& \textbf{90.8}\\
		\textbf{PR5}  &$F_1$(\%)    & 89.7 & 54.9 & 89.7 & 75.9 & 68.2 & 78.4 & 90.4 &80.7& \textbf{92.3}\\
		  &               $AUC$(\%)    & 86.4 & 72.8 & 85.9 & 90.4 & 64.8 & 82.7  & 87.9 &91.5& \textbf{97.4}\\
    \bottomrule[2pt]
	\end{tabular}
	}
\end{table*}

\begin{table*}[ht]
	\centering
	\caption{\centering{PERFORMANCE COMPARISON OF STA-GT AND ALL BASELINES ON THE PUBLIC DATASET.}}
	\label{tab:3}   
	\scalebox{1}{
	\begin{tabular}{l|c|ccccccccc}
		\toprule[2pt]
		\textbf{Dataset}     & \textbf{Criteria}     &  \textbf{GCN}    & \textbf{GraphSAGE} & \textbf{GAT}  &\textbf{CARE-GNN}   &\textbf{SSA}    &\textbf{RGCN} &\textbf{HAN}   & \textbf{FRAUDRE}  & \textbf{STA-GT}\\
            \midrule

		  &               $Recall$(\%) & 57.4  & 55.3  & 58.9    &  62.8  &  58.7 &53.5&56.6& 62.7 &\textbf{72.7}\\
		\textbf{TC12}  &$F_1$(\%)    & 67.8 & 63.8 & 66.9   &  70.2  &  69.7 &66.1&61.2&56.9  &\textbf{71.1}\\
		  &               $AUC$(\%)   & 56.9 & 61.4 & 77.1    &  68.8  &  75.8 &65.2&72.9&81.4 &\textbf{89.7}\\
            \midrule
    		  &           $Recall$(\%) & 26.8 & 51.6 & 66.6 & 62.1&59.1 &52.1&63.5&77.6& \textbf{81.9}\\
		\textbf{TC23}  &$F_1$(\%)    & 42.3 & 67.4  & 66.6   &   69.6  & 70.4 &67.6&69.1&78.2&\textbf{78.8}\\
		  &               $AUC$(\%)    & 76.0 & 75.5 & 75.9    &  66.6 &  60.4 &63.7&59.7&84.8&\textbf{89.9}\\
            \midrule
            &               $Recall$(\%) & 47.0 & 50.6 & 62.9    & 66.5    &63.8 &56.7&51.3&66.3&\textbf{81.5}\\
		\textbf{TC34}  &$F_1$(\%)    & 61.6 & 67.2  & 62.8   &  63.7   & 67.9 &72.4&67.8&60.5 &\textbf{82.0}\\
		  &               $AUC$(\%)    & 58.2 & 51.2  & 71.5    &  74.1    &72.8 &54.5&63.0&77.6& \textbf{84.3}\\
    \bottomrule[2pt]
	\end{tabular}
	}
\end{table*}
\subsection{Performance Comparison}

We conduct a comparative analysis between the proposed method STA-GT and the baseline models on two financial datasets. The results are reported in Tables.~I and II. We have the following observations.

Compared to GCN, GraphSAGE, and GAT modeled on the graph with a single relation, RGCN and HAN running on the multi-relation graph did not perform better on the two datasets. The reason is that directly employing GNN models for the identification of fraudulent transactions is unsuitable. While the utilization of multi-graphs offers a broader range of information and more complex relationships, it is crucial to handle node interactions with caution and avoid introducing dissimilarity information, ensuring the opportunity for enhanced performance. FRAUDRE has achieved promising performance by introducing the fraud-aware module and an imbalance-oriented loss function to tackle graph inconsistency and imbalance issues.

As shown in Tables.~I and II, the proposed method STA-GT outperforms all baselines with at least $3.8\%$, $1.1\%$, $1.7\%$, $5.1\%$, $5.2\%$, $9.9\%$, $4.3\%$, and $15.0\%$ $Recall$ improvements on all datasets.
Meanwhile, STA-GT outperforms the other baselines with at least $2.7\%$, $9.9\%$, $1.3\%$, $3.4\%$, $11.6\%$, $3.3\%$, $0.6\%$, and $14.1\%$ $F_1$ improvements. The $AUC$ score of our method also improved on most datasets. These experimental results provide strong evidence of the superiority of STA-GT for the identification of fraudulent transactions.

\section{Conclusion}
In this paper, we propose a novel heterogeneous graph neural network framework called STA-GT to tackle the transaction fraud detection problem. To integrate spatial-temporal information and enlarge the receptive field, we design the temporal encoding strategy and combine it with heterogeneous graph convolution operation to learn node representations. Furthermore, a transformer module is built on the top of the above GNN layer stack to learn global and local information jointly. It utilizes informative but long-distance transaction records effectively, which can ensure both intraclass compactness and interclass separation.
Experimental results on two financial datasets show the superiority of STA-GT on the transaction fraud detection task. In the subsequent work, we will explore how the explainability of the GNN model.

\bibliographystyle{IEEEtran}

\bibliography{arxiv_TII}


 





\end{document}